\documentclass[10pt,journal,compsoc]{IEEEtran}
\usepackage{bm}
\usepackage{float}
\usepackage{color}
\usepackage{array}
\usepackage{xcolor}
\usepackage{amssymb}
\usepackage{amsmath}
\usepackage{stfloats}
\usepackage{subfiles}
\usepackage{ragged2e}
\usepackage{colortbl}
\usepackage{multirow}
\usepackage{graphicx}
\usepackage{booktabs}
\usepackage{subfigure}
\usepackage{longtable}
\usepackage[switch]{lineno}
\usepackage[justification=justified]{caption}
\usepackage[square, comma, sort&compress, numbers]{natbib}
\usepackage[colorlinks, linkcolor=red, anchorcolor=red, citecolor=green,urlcolor=blue]{hyperref}

\definecolor{mark}{rgb}{0,0,0}
\definecolor{table}{rgb}{0,0,0}

\usepackage{algorithm}
\usepackage{algpseudocode}

\begin{document}
\title{A Teacher-Free Graph Knowledge Distillation Framework with Dual Self-Distillation}

\author{Lirong~Wu,~Haitao~Lin,~Zhangyang~Gao,~Guojiang~Zhao,~and~Stan Z. Li$^\dagger$,~\IEEEmembership{Fellow,~IEEE}
        
\IEEEcompsocitemizethanks{
\IEEEcompsocthanksitem Lirong Wu, Haitao Lin, Zhangyang Gao, Guojiang Zhao, and Stan Z. Li are with the AI Lab, Research Center for Industries of the Future, Westlake University, Hangzhou 310000, China. E-mail: \{wulirong, linhaitao, gaozhangyang, zhaoguojiang, stan.zq.li\}@westlake.edu.cn. \\ $\dagger$ Corresponding Author.}}


\IEEEtitleabstractindextext{
\justifying
\begin{abstract}
Recent years have witnessed great success in handling graph-related tasks with Graph Neural Networks (GNNs). Despite their great \emph{academic} success, Multi-Layer Perceptrons (MLPs) remain the primary workhorse for practical \emph{industrial} applications. One reason for such an academic-industry gap is the neighborhood-fetching latency incurred by data dependency in GNNs. To reduce their gaps, Graph Knowledge Distillation (GKD) is proposed, usually based on a standard teacher-student architecture, to distill knowledge from a large teacher GNN into a lightweight student GNN or MLP. However, we found in this paper that neither teachers nor GNNs are necessary for graph knowledge distillation. We propose a \emph{\underline{T}eacher-Free \underline{G}raph \underline{S}elf-Distillation} (TGS) framework that does not require any teacher model or GNNs during both training and inference. More importantly, the proposed TGS framework is purely based on MLPs, where structural information is only implicitly used to guide \emph{dual knowledge self-distillation} between the target node and its neighborhood. As a result, TGS enjoys the benefits of graph topology awareness in training but is free from data dependency in inference. Extensive experiments have shown that the performance of vanilla MLPs can be greatly improved with dual self-distillation, e.g., TGS improves over vanilla MLPs by 15.54\% on average and outperforms state-of-the-art GKD algorithms on six real-world datasets. In terms of inference speed, TGS infers 75$\times$-89$\times$ faster than existing GNNs and 16$\times$-25$\times$ faster than classical inference acceleration methods. 
\end{abstract}

}

\maketitle

\section{Introduction}
Recently, Graph Neural Networks (GNNs) \cite{wu2020comprehensive,zhou2020graph,wu2023homophily,wu2022graphmixup,wu2023beyond} have demonstrated their powerful capability to handle graph-structured data in a variety of applications, including meteorology \cite{lin2022conditional,ma2023histgnn}, life sciences \cite{wu2024mapeppi,wu2024protein}, and transportation \cite{zhou2020variational,jiang2022graph}. Despite their great academic success, practical deployments of GNNs in the industry are still less popular. One reason for their gaps is the neighborhood-fetching inference latency incurred by data dependency in GNNs \cite{jia2020redundancy}. Most existing GNNs \cite{kipf2016variational,velivckovic2017graph,hamilton2017inductive} rely on message passing to \textcolor{mark}{aggregate neighborhood features for capturing data dependency between nodes within a local subgraph}. As a result, neighborhood fetching caused by data dependency is one of the major sources of GNN latency during inference. For example, to infer a single node with a $L$-layer GNN on a graph with average node degree $R$, it requires fetching and \textcolor{mark}{aggregating $\mathcal{O}(R^L)$ fetched nodes}. However, $R$ can be large for real-world graphs, e.g., 19 for the Amazon-com dataset, and $L$ is getting deeper for the latest GNN architectures, e.g., $L$=1001 layers for RevGNN-Deep \cite{li2021training}. Compared with GNNs, MLPs do not suffer from the data dependency problem and infer much faster than GNNs. However, due to the lack of modeling data dependencies, MLPs may fail to take full advantage of the graph topological information, which limits their performance on various downstream tasks.

\begin{figure*}[!tbp]
	\begin{center}
  		\subfigure[Four different types of GKD algorithms]{\includegraphics[width=0.43\linewidth]{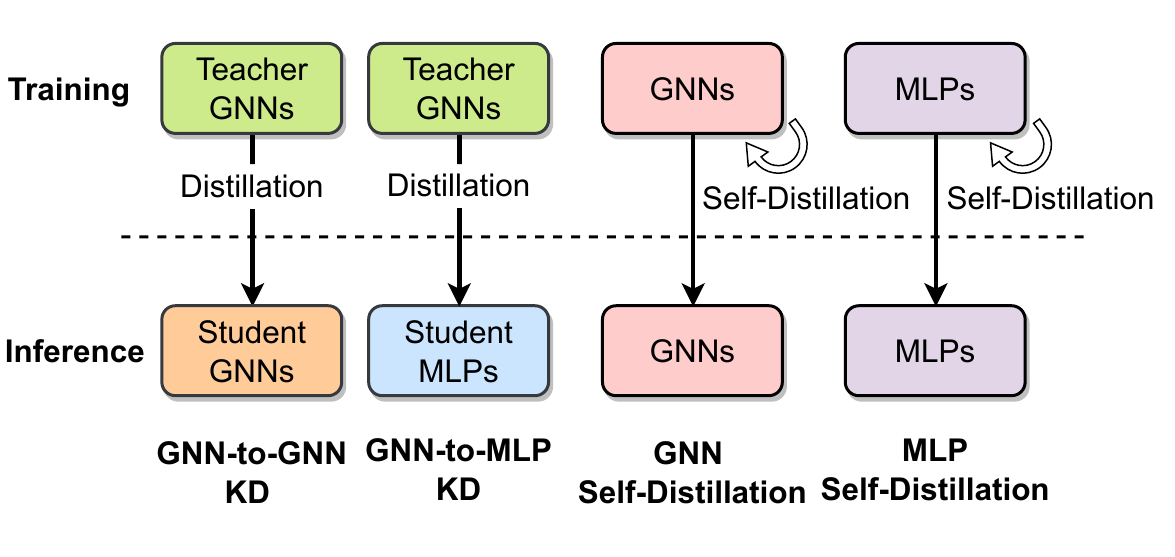} \label{fig:1b}}
		\subfigure[Inference Accuracy \textit{vs.} Inference Time]{\includegraphics[width=0.30\linewidth]{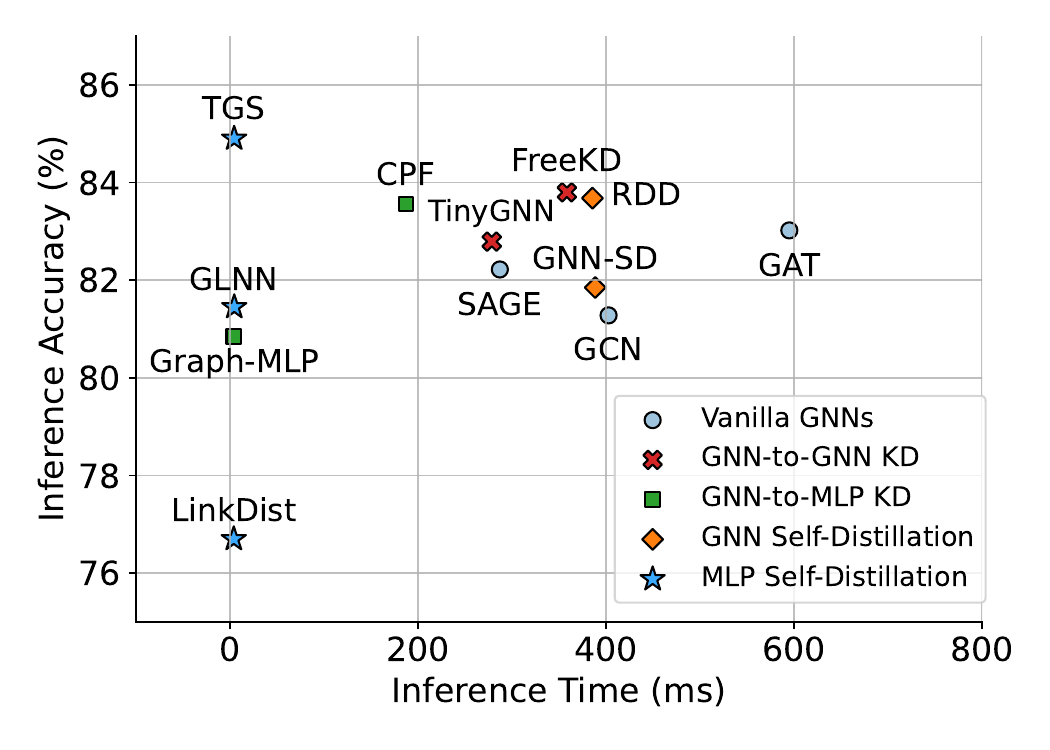} \label{fig:1c}}
	\end{center}
	\caption{(a): Illustration of four different types of graph knowledge distillation algorithms, depending on whether teacher models and GNNs/MLPs are included in training and inference. (b): Inference accuracy (\%) \textit{vs.} inference time (ms) on the Cora dataset. If not specifically mentioned, all GKD algorithms adopt GCN as the backbone by default.}
	\label{fig:1}
\end{figure*}

To connect the two worlds of topology-aware GNNs and inference-efficient MLPs, graph knowledge distillation \cite{yan2020tinygnn,he2022compressing,zhang2023iterative,nardini2022distilled,liu2019learning} is proposed to distill knowledge from large teacher GNNs to small student GNNs or MLPs, yielding two branches, namely GNN-to-GNN \cite{yang2020distilling,ren2021multi,wu2022knowledge} and GNN-to-MLP \cite{zhang2021graph,yang2021extract,wu2023extracting,anonymous2023double,wu2023quantifying}, as shown in Fig.~\ref{fig:1b}. Different from the standard teacher-student KD, there is a similar technique called GNN Self-Distillation \cite{chen2020self,zhang2020reliable}, which regularizes a GNN model by \emph {distilling its own knowledge within GNNs (e.g., across different layers or nodes) without other teacher models}. Similarly, MLP Self-Distillation \cite{hu2021graph,luo2021distilling} is proposed to implicitly self-distill knowledge from structural information, but it does not involve any teachers or GNNs and is purely based on MLPs.

Considering both inference accuracy and inference time, the above four types of graph knowledge distillation methods are two completely different worlds. As shown in Fig.~\ref{fig:1c}, GNN-to-GNN KD and GNN Self-Distillation directly deploy well-distilled GNNs for inference, which helps to improve the inference accuracy but does not help much in inference acceleration with the data dependency in GNNs unresolved. On the other hand, GNN-to-MLP KD and MLP Self-Distillation require only MLPs for inference, which greatly improves their inference speed, but brings limited performance improvement. In particular, MLP Self-Distillation completely removes GNNs, which disables it from being topology-aware, making its inference fastest, but at the cost of undesired performance drops.

In this paper, we found that neither teachers nor GNNs are necessary to achieve \textbf{high-accuracy} and \textbf{high-efficiency} inference on graphs. Therefore, we propose a simple \emph{\underline{T}eacher-Free \underline{G}raph \underline{S}elf-Distillation} (TGS) framework that does not require any teacher model or GNNs during both training and inference. The proposed TGS framework is purely based on MLPs, where structural information is only implicitly used to guide \emph{dual knowledge self-distillation} between the target node and its neighborhood, substituting the explicit information propagation as in GNNs. As a result, the resulting model enjoys the benefits of graph topology-awareness in training but reduces time overhead in inference. Moreover, we propose an edge sampling strategy for batch-style self-distillation instead of feeding the entire graph into memory to reduce memory usage when scaling to large-scale graphs. Extensive experiments show that TGS infers as fast as MLPs, but its inference accuracy is comparable to or even better than state-of-the-art GKD algorithms. We believe that this work will inspire researchers to rethink the necessity of teachers and GNNs for graph knowledge distillation. Codes will be public at \url{https://github.com/LirongWu/TGS}.

\section{Related Work}
\subsection{Graph Neural Networks}
Graph Neural Networks (GNNs) can be mainly divided into two categories, i.e., spectral-based GNNs and spatial-based GNNs. The spectral-based GNNs, such as ChebyNet \cite{defferrard2016convolutional} and GCN \cite{kipf2016semi}, define graph convolution kernels in the spectral domain based on the graph signal processing theory. Instead, the spatial-based GNNs, such as GraphSAGE \cite{hamilton2017inductive} and GAT \cite{kipf2016semi}, directly define updating rules in the spatial space and focus on the design of neighborhood aggregation functions. The closest work to ours is APPNP \cite{klicpera2018predict}, which shares some similar design insights with TGS in that both first use graph-independent models to extract node embeddings. However, their main difference is that APPNP applies Personalized PageRank to \textbf{explicitly} encode the adjacency matrix via message passing, while TGS uses Mixup and two inference layers to \textbf{implicitly} exploit the adjacency matrix as supervision signals. We refer interested readers to the recent survey \cite{wu2020comprehensive,zhang2020deep} for more GNN architectures. Despite their great progress, the above GNNs all share the de facto design that structural information is explicitly used for message passing, which leaves neighborhood fetching still one major source of GNN inference latency. To reduce multiplication and accumulation operations, many \textbf{inference acceleration} technologies have been applied, including Pruning \cite{han2015learning}, Quantizing \cite{gupta2015deep}, and Neighborhood Sampling \cite{chen2018fastgcn}.

\subsection{Graph Knowledge Distillation}
Several previous works on graph distillation try to distill knowledge from large teacher GNNs to smaller student GNNs, termed as \emph{GNN-to-GNN KD} \cite{xu2022ccgl}. The student model in FreeKD \cite{feng2022freekd} and TinyGNN \cite{yan2020tinygnn} is a GNN with fewer parameters, but not necessarily fewer layers, which makes them still suffer from the neighborhood-fetching latency. The other branch of graph knowledge distillation is to distill from large teacher GNNs to lightweight student MLPs, termed as \emph{GNN-to-MLP KD}. For example, CPF \cite{yang2021extract} proposes to distill knowledge from teacher GNNs to student MLPs, but it takes advantage of label propagation \cite{iscen2019label} in MLPs to improve performance and thus remains heavily data-dependent. Besides, GLNN \cite{zhang2021graph} directly distills knowledge from GNNs to student MLPs, which has a great advantage in inference speed, but its performance gains are limited. 

Different from the standard teacher-student KD architecture, there is a similar technique known as \emph{GNN Self-Distillation} \cite{zhang2019your}, which regularizes a GNN model by distilling its own knowledge without other teacher models. For example, GNN-SD \cite{chen2020self} directly self-distills knowledge across different GNN layers. Besides, RDD \cite{zhang2020reliable} builds an ensemble teacher using multiple versions of the model itself without introducing an external teacher model. Compared to GNNs, MLPs have no data dependency and infer much faster. There are some attempts on \emph{MLP Self-Distillation}, such as Graph-MLP \cite{hu2021graph} and LinkDist \cite{luo2021distilling}, which are purely based on MLP and self-distill knowledge from structural information by contrastive or consistency constraints. Despite the great advantage of MLP Self-Distillation in inference speed, their inference accuracy still cannot match the state-of-the-art competitors. More related work on graph knowledge distillation can be found in a recent survey \cite{tian2023knowledge}. The closest work to ours is Graph-MLP, but the essential difference between Graph-MLP and TGS is that the former is a contrastive learning method, while the latter is based on knowledge distillation. Their main differences are in the following four aspects: (1) TGS utilizes mixup as data augmentation, while Graph-MLP \cite{hu2021graph}, as a contrastive learning method, instead discards this; (2) Graph-MLP treats all nodes within an $r$-hop neighborhood as positive samples with $\mathcal{O}(R^r)$ complexity, whereas TGS performs distillation only within a 1-hop neighborhood with a complexity of $\mathcal{O}(R)$, where $R$ is the average node degree; (3) Graph-MLP treats all nodes within a batch as negative samples, while TGS only takes one randomly selected node as negative sample; and (4) as a contrastive learning method, Graph-MLP relies on InfoNCE to perform contrasting, whereas TGS directly minimizes the MSE losses between samples. 

\subsection{Graph Contrastive Learning (GCL)}
Recent years have witnessed the great success of graph contrastive learning in learning graph representation \cite{wu2021self}. Graph contrastive learning generates multiple views for each instance through various data augmentation methods and maximizes the agreement between two positive samples (as measured by mutual information) against a large number of negative samples. For example, Deep Graph Infomax (DGI) \cite{velickovic2019deep} is proposed to contrast the patch representations and corresponding high-level summary of graphs. GraphCL \cite{you2020graph} is one of the pioneering works that extends contrastive learning to graphs, and it defines four graph augmentation methods that achieve good results on the molecular property prediction task. Its follow-up, GCA \cite{zhu2020graph}, combines adaptive augmentation, which further improves the performance. Furthermore, InfoGCL \cite{xu2021infogcl} investigates how information is transformed and transferred during contrastive learning based on the information bottleneck, aimed at minimizing the loss of task-relevant information. \textcolor{mark}{There has been some recent work exploring the necessity of negative samples for graph contrastive learning. Inspired by BYOL \cite{grill2020bootstrap}, BGRL \cite{thakoor2021bootstrapped} proposes to perform the self-supervised learning that \emph{does not require negative samples}, thus getting rid of the potentially quadratic bottleneck. Moreover, in addition to the commonly used InfoNCE \cite{gutmann2010noise} as a loss function for measuring agreement, MSE is also widely used in GCL, such as SelfGNN \cite{kefato2021self}, DMGI \cite{park2020unsupervised}, etc.}

\section{Preliminaries}
\textbf{Notations and Problem Statement.}
Let $\mathcal{G}=(\mathcal{V}, \mathcal{E})$ denote a graph, where $\mathcal{V}$ is the set of $|\mathcal{V}|=N$ nodes with features $\mathbf{X}=\left[\mathbf{x}_{1}, \mathbf{x}_{2}, \cdots, \mathbf{x}_{N}\right]\in \mathbb{R}^{N \times d}$ and $\mathcal{E}$ is the set of edges between nodes. Each node $v_i \in \mathcal{V}$ is associated with a $d$-dimensional features vector $\mathbf{x}_{i}$. Following the common semi-supervised node classification setting, only a subset of node $\mathcal{V}_L=\{v_1,v_2,\cdots,v_L\}$ with corresponding labels $\mathcal{Y}_L=\{y_1,y_2,\cdots,y_L\}$ are known, and we denote the labeled set as $\mathcal{D}_L=(\mathcal{V}_L,\mathcal{Y}_L)$ and unlabeled set as $\mathcal{D}_U=(\mathcal{V}_U,\mathcal{Y}_U)$, where $\mathcal{V}_U=\mathcal{V} \backslash \mathcal{V}_L$. The task of node classification aims to learn a mapping $\Phi: \mathcal{V}\!\rightarrow\!\mathcal{Y}$ on labeled data $\mathcal{D}_L$, so that it can be used to infer the labels $\mathcal{Y}_U$ \cite{yang2022survey}.
\newline

\noindent \textbf{Graph Neural Networks.}
A general GNN framework consists of two key computations for each node $v_i$ at every layer: (1) $\operatorname{AGGREGATE}$ operation: aggregating messages from neighborhood $\mathcal{N}_i$; (2) $\operatorname{UPDATE}$ operation: updating node representation from its representation in the previous layer and aggregated messages. Considering a $L$-layer GNN, the formulation of the $l$-th layer is as follows,
\begin{equation}
\begin{aligned}
\mathbf{m}_{i}^{(l)} & = \operatorname{AGGREGATE}^{(l)}\left(\left\{\mathbf{h}_{j}^{(l-1)}: v_{j} \in \mathcal{N}_i\right\}\right) \\
\mathbf{h}_{i}^{(l)} & = \operatorname{UPDATE}^{(l)}\left(\mathbf{h}_{i}^{(l-1)}, \mathbf{m}_{i}^{(l)}\right)
\end{aligned}
\label{equ:1}
\end{equation}
where $0\leq l \leq L-1$, $\mathbf{h}_{i}^{(l)}$ is the embedding of node $v_i$ in the $l$-th layer, and $\mathbf{h}_{i}^{(0)}\!=\!\mathbf{x}_{i}$ is the input feature. After $L$ message-passing layers, the final node embeddings $\mathbf{h}_{i}^{(L)}$ can be passed through an additional linear inference layer $\mathbf{y}_i=f_\theta(\mathbf{h}_i^{(L)})$ for node classification on the target node $v_i$.

\section{Methodology}
\begin{figure*}[!htbp]
	\begin{center}
		\includegraphics[width=1.0\linewidth]{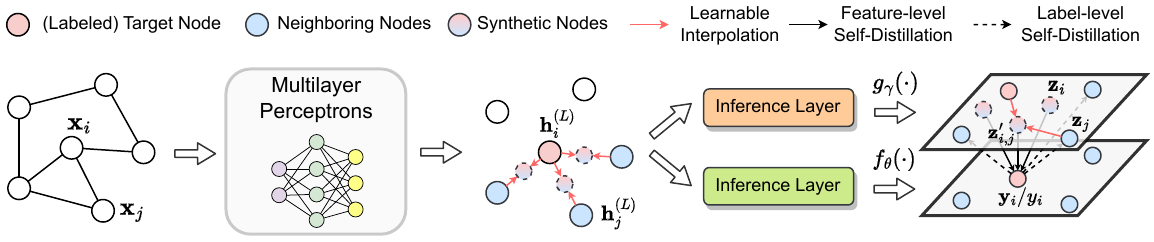}
	\end{center}
	\caption{Illustration of the proposed \emph{Teacher-Free Graph Self-Distillation} (TGS) framework. In the training stage, the MLP and two inference layers $f_\theta(\cdot)$, $g_\gamma(\cdot)$ are jointly trained by the proposed dual feature-level and label-level self-distillation.}
	\label{fig:2}
\end{figure*}

Motivated by the complementary strengths and weaknesses of GNNs and MLPs, we propose in this paper a simple but effective \emph{\underline{T}eacher-Free \underline{G}raph \underline{S}elf-Distillation} (TGS) framework, as illustrated in Fig.~\ref{fig:2}. The proposed TGS framework enjoys the benefits of GNN-like topology-awareness in training but keeps the inference-efficiency of MLPs in inference. The following subsections focus on three key aspects: (1) backbone architecture, how to construct a ``boosted" MLP as backbone; (2) dual self-distillation, how to self-distill knowledge between the target node and its neighborhood; (3) training and inference, how to solve training difficulties and make inference with the trained model.

\subsection{Backbone Architecture}
The TGS framework is based on pure MLPs, with each layer composed of a linear transformation, an activation function, a batch normalization, and a dropout function, defined as:
\begin{equation}
\mathbf{H}^{(l+1)}=\operatorname{Dropout}\big(B N\big(\sigma\big(\mathbf{H}^{(l)} \mathbf{W}^{(l)}\big)\big)\big),\quad \mathbf{H}^{(0)}=\mathbf{X} 
\end{equation}
where $0 \leq l \leq L-1$, $\sigma=\mathrm{ReLu}(\cdot)$ denotes an activation function, $BN(\cdot)$ denotes the batch normalization, and $\operatorname{Dropout}(\cdot)$ is the dropout function. In addition, $\mathbf{W}^{(0)} \in \mathbb{R}^{d \times F}$ and $\mathbf{W}^{(l)} \in \mathbb{R}^{F \times F}$ $(1 \leq l \leq L-1)$ are layer-specific parameter matrices with the hidden dimension $F$. 

Given a target node $v_i$ and its neighboring nodes $\mathcal{N}_i$, we first feed their node features $\mathbf{x}_i$ and $\{\mathbf{x}_j\}_{j\in\mathcal{N}_i}$ into MLPs and encode them as hidden representations $\mathbf{h}^{(L)}_i$ and $\{\mathbf{h}^{(L)}_j\}_{j\in\mathcal{N}_i}$. Then, we define two parameter-independent inference layers for label prediction, respectively, i.e., $\mathbf{y}_i=f_\theta(\mathbf{h}_i^{(L)}) \in \mathbb{R}^{C}$ and $\mathbf{z}_j=g_\gamma(\mathbf{h}_j^{(L)}) \in \mathbb{R}^{C}$, where $C$ is the number of category. Both two inference layers are implemented with one layer of linear transformation by default in this paper. In the next subsections, we will discuss in detail how to implicitly self-distill graph knowledge between the target node $v_i$ and its neighborhood nodes $\mathcal{N}_i$.

\subsection{Dual Knowledge Self-Distillation}
In this subsection, we propose dual knowledge self-distillation to learn node features $\mathbf{X}$ and labels $\mathcal{Y}_L$ guided by structural information $\mathcal{E}$. It consists of two \textbf{bidirectional} modules, one \emph{Feature-level Self-Distillation} that distills feature information from the neighborhood into the target node, and one \emph{Label-level Self-Distillation} that distills label information from the target node into the neighborhood.

\subsubsection{Feature-level Self-Distillation}
The graphs can be categorized into homophily and heterophily graphs, where the former fulfills the smoothness assumption while the latter does not. Considering the importance and prevalence of homophily graphs, for which a variety of classical GNNs including GCN, GAT, SAGE, and APPNP have been previously developed, we focus mainly on learning homophily graphs in this paper. The smoothness assumption indicates that neighboring nodes in a graph tend to share similar features and labels, while non-neighboring nodes should be far away. With such motivation, we perform feature-level self-distillation on the neighborhood by regularizing the consistency of the label distributions between the target node and its neighboring nodes. Along with connectivity, disconnectivity between nodes also carries important information that reveals the node's dissimilarity. However, the number of neighboring nodes is much smaller compared with those non-neighboring nodes, which renders the model overemphasize the differences between the target and non-neighboring nodes, possibly leading to imprecise class boundaries. To solve this problem, we modify \emph{Mixup} \cite{zhang2017mixup}, an effective data augmentation that performs interpolation between samples to generate new samples, to augment neighboring nodes. Specifically, we perform \textit{learnable interpolation} between the target node $v_i$ and its neighboring node $v_j\in\mathcal{N}_i$ to generate a new node, with its node representation defined as
\begin{equation}
\begin{aligned}
\mathbf{z}^\prime_{i,j} = &  g_\gamma\big(\beta_{i,j}\mathbf{h}_j^{(L)}+(1-\beta_{i,j})\mathbf{h}_i^{(L)}\big), \\ \text{where}  & \ \ \  \beta_{i,j}=\mathrm{sigmoid}\big(\mathbf{a}^T \big[\mathbf{x}_i\mathbf{W}_m\|\mathbf{x}_j\mathbf{W}_m\big]\big)
\label{equ:3}
\end{aligned}
\end{equation}
where $\mathbf{W}_m \in \mathbb{R}^{d \times F}$ and $\beta_{i,j}$ is defined as \textit{learnable interpolation coefficients} with the shared attention weight $\mathbf{a}$. Then, we take augmented neighboring nodes as positive samples and other non-neighboring nodes as negative samples to simultaneously model the connectivity and disconnectivity between nodes. Specifically, the learning objective of feature-level self-distillation is defined as follows,
\begin{equation}
\begin{small}
\begin{aligned}
\mathcal{L}_{\text{feat}}=\frac{1}{N}\sum_{i=1}^N\Big(\frac{1}{|\mathcal{N}_i|}\sum_{j\in\mathcal{N}_i} \big\|\mathbf{y}_i-\mathbf{z}^\prime_{i,j}\big\|_2^2-\frac{1}{M_i}\sum_{e_{i,k}\notin\mathcal{E}} \big\|\widehat{\mathbf{y}}_i-\widehat{\mathbf{z}}_k\big\|_2^2\Big)
\end{aligned}
\end{small}
\label{equ:4}
\end{equation}
where $M_i\!=\!|\mathcal{E}|\!-\!|\mathcal{N}_i|\!-\!1$ is the number of negative samples of node $v_i$. Besides, $\widehat{\mathbf{y}}_{i}=\textrm{softmax}(\mathbf{y}_i) \in \mathbb{R}^C$ and $\widehat{\mathbf{z}}_{k}=\textrm{softmax}(\mathbf{z}_k) \in \mathbb{R}^C$. The loss function $\mathcal{L}_{\text{feat}}$ defined in Eq.~(\ref{equ:4}) essentially encourages positive neighboring nodes to be closer and pushes negative non-neighboring nodes away.

\subsubsection{Label-level Self-Distillation}
Thus far, we have only discussed how to use node features $\mathbf{X}$ and structural information $\mathcal{E}$ for feature-level self-distillation, but have not explored how to leverage node labels $\mathcal{Y}_L$. A widely used solution to leverage labels is to optimize the objective on the labeled data $\mathcal{V}_L$ as follows
\begin{equation}
\min_{\theta,\mathbf{W}^{(0)},\cdots,\mathbf{W}^{(L-1)}} \frac{1}{|\mathcal{V}_L|}\sum_{i\in\mathcal{V}_L} \mathcal{L}_{CE}(y_i, \widehat{\mathbf{y}}_i),
\label{equ:5}
\end{equation}
where $\mathcal{L}_{CE}(y_i, \widehat{\mathbf{y}}_i)$ denotes the cross-entropy loss between $\widehat{\mathbf{y}}_{i}$ and ground-truth label $y_i$. However, Eq.~(\ref{equ:5}) only considers node labels $\mathcal{Y}_L$, but completely ignores structural information $\mathcal{E}$. In practice, label propagation \cite{zhu2002learning} is widely used as an effective trick to simultaneously model label and structural information and achieves promising results for various GNNs. However, label propagation involves the explicit coupling of labels with the structure, so it is heavily data-dependent with the same inference-latency problem as message passing. Earlier, we have proposed feature-level self-distillation of Eq.~(\ref{equ:4}) to substitute explicit message passing in Eq.~(\ref{equ:1}), and next we introduce \textit{implicit} label-level self-distillation of Eq.~(\ref{equ:6}) to substitute \textit{explicit} label propagation to jointly exploit both label and structural information. The objective of label-level self-distillation is defined as
\begin{equation}
\mathcal{L}_{\text{label}} = \frac{1}{|\mathcal{V}_L|}\sum_{i\in\mathcal{V}_L} \Big(\mathcal{L}_{CE}(y_i, \widehat{\mathbf{y}}_i) + \sum_{j\in\mathcal{N}_i} \mathcal{L}_{CE}(y_i, \widehat{\mathbf{z}}_j)\Big).
\label{equ:6}
\end{equation}

\subsection{Training and Inferring}
\subsubsection{Model Training}
In practice, directly optimizing Eq.~(\ref{equ:4})(\ref{equ:6}) faces two tricky challenges: \textit{(1)} it treats all non-neighboring nodes as negative samples, which suffers from a huge computational burden; and \textit{(2)} it performs the summation over the entire set of nodes, i.e, requiring a large memory space for keeping the entire graph. To address these two problems, we adopt the edge sampling strategy \cite{mikolov2013distributed,tang2015line} instead of feeding the entire graph into the memory for \emph{batch-style training}. More specifically, we first sample mini-batch edges $\mathcal{E}_b\in\mathcal{E}$ from the entire edge set. Then we randomly sample negative nodes by a pre-defined negative distribution $P_k(v)$ for each edge $e_{i,j}\in\mathcal{E}_b$ instead of enumerating all non-neighboring nodes as negative samples. Finally, we can rewrite Eq.~(\ref{equ:4}) as
\begin{equation}
\begin{small}
\begin{aligned}
\mathcal{L}_{\text{feat}}  \! =\! \frac{1}{B}\sum_{b=1}^B & \sum_{e_{i,j}\in\mathcal{E}_b}  \bigg(
\big\|\mathbf{y}_i-\mathbf{z}^\prime_{i,j}\big\|_2^2
+\big\|\mathbf{y}_j-\mathbf{z}^\prime_{j,i}\big\|_2^2 \\
&-\mathbb{E}_{{v_k} \sim P_k(v)} \Big(
\big\|\widehat{\mathbf{y}}_i-\widehat{\mathbf{z}}_k\big\|_2^2
+\big\|\widehat{\mathbf{y}}_j-\widehat{\mathbf{z}}_k\big\|_2^2\Big)
\bigg),
\end{aligned}
\end{small}
\label{equ:7}
\end{equation}
where $B$ is the batch size, and $P_k(v)$ adopts the uniform distribution by default, that is $P_k(v_i)=\frac{1}{N}$ for each node $v_i$. $P_k(v)$ can also be pre-defined based on prior knowledge, e.g., degree distribution, but in practice, we find from the experimental results in Sec.~\ref{sec:5.5} that uniform distribution is a reasonable choice that can yield fairly good performance across various datasets. Similarly, we can rewrite the label-level self-distillation in Eq.~(\ref{equ:6}) as a batch-style formulation,
\begin{equation}
\hspace{-1em}
\begin{small}
\begin{aligned}
\mathcal{L}_{\text{label}}=\frac{1}{B}\sum_{b=1}^B \frac{1}{|\mathcal{V}_b|}\sum_{i\in\mathcal{V}_b}
\Big(\mathcal{L}_{CE}(y_i, \widehat{\mathbf{y}}_i) + \sum_{e_{i,j}\in\mathcal{E}_b} \mathcal{L}_{CE}(y_i, \widehat{\mathbf{z}}_j)\Big),
\end{aligned}
\end{small}
\label{equ:8}
\end{equation}
where $\mathcal{V}_b=\{v_i,v_j|e_{i,j}\in\mathcal{E}_b\}\cap\mathcal{V}_L$ is all the sampled nodes in $\mathcal{E}_b$. Finally, the total training loss can be defined as, 
\begin{equation}\mathcal{L}_{\text{total}}=\mathcal{L}_{\text{label}}+\alpha\mathcal{L}_{\text{feat}},
\label{equ:9}
\end{equation} 
where $\alpha$ is a trade-off hyperparameter.

\subsubsection{Model Inferring}
Once the model training is completed, we can directly omit the inference layer $g_\gamma(\cdot)$ and retain the backbone MLP architecture and the inference layer $f_\theta(\cdot)$ for label prediction. At this time, there is no data dependency for model inference, and this is attributed to the fact that we have shifted a considerable amount of work from the latency-sensitive inference stage to the latency-insensitive training stage. The pseudo-code of TGS is summarized in Algorithm.~\ref{algo:1}.

\begin{algorithm}[!htbp]
	\caption{Algorithm for the proposed \textit{TGS} framework}
	\label{algo:1}
	\begin{algorithmic}[1]
		\Require Features: $\mathbf{X}$; Edge Set: $\mathcal{E}$; \# Batch: $B$; \# Epoch: $E$. 
		
		\State Initialize parameters $\{\mathbf{W}^{l}\}_{l=0}^{L-1}$, $f_\theta(\cdot)$, and $g_\gamma(\cdot)$.
		
		\For{$epoch$ $\in$ \{0,1,$\cdots$,$E-1$\}}
		    \For{$b$ $\in$ \{0,1,$\cdots$,$B-1$\}}
		        \State Sample a mini-batch of edges $\mathcal{E}_b$ from $\mathcal{E}$;
		        \State Compute losses $\mathcal{L}_{\text{feat}}$ and $\mathcal{L}_{\text{label}}$ by Eq.~(\ref{equ:7})(\ref{equ:8});
		        \State Sum up $\mathcal{L}_{\text{feat}}$ and $\mathcal{L}_{\text{label}}$ as total loss $L_{\text{total}}$;
		        \State Update parameters by back-propagation of $L_{\text{total}}$.
		    \EndFor
		\EndFor
        \State Predict labels $\mathcal{Y}_U$ for those unlabeled nodes $\mathcal{V}_U$.
		\State \textbf{return} Predictions $\mathcal{Y}_U$, parameters $\{\mathbf{W}^{l}\}_{l=0}^{L-1}$ and $f_\theta(\cdot)$.
	\end{algorithmic}
\end{algorithm}

\subsection{Discussion and Comparison}
\textbf{Comparison with Message and Label Propagation.}
The core of GNNs is the use of structural information, where \emph{message passing} and \emph{label propagation} are the two dominant schemes. \textcolor{mark}{The message passing models the data dependency within a local subgraph through neighborhood feature aggregation}, while label propagation focuses on diffusing label information to the neighborhood, and \emph{they are complementary to each other}. However, both of them involve explicit coupling of features/labels with structures, leading to data dependency and inference latency. Different from the \textit{explicit} message passing and label propagation, we use structural information as prior, as shown in Fig.~\ref{fig:1}, to \textit{implicitly} guide bidirectional dual knowledge self-distillation: (1) feature-level self-distillation from neighborhood to the target node as in Eq.~(\ref{equ:4}) and (2) label-level self-distillation from the target node to the neighborhood as in Eq.~(\ref{equ:6}), where structural information is never explicitly involved in the forward propagation. Furthermore, while Eq.~(\ref{equ:4})(\ref{equ:6}) is defined on the 1-hop neighborhood, it can aggregate messages from multi-hops away by multiple cascades of self-distillation. For example, if there is knowledge self-distillation between the target node and its 1-hop neighbor and between its 1-hop neighbor and its 2-hop neighbor, then as the training proceeds, the messages from the 2-hop neighbor will first be distilled to the 1-hop neighbor, and then progressively propagated to the target node in such a "cascading" manner.
\newline

\noindent \textbf{Comparison with Graph Contrastive Learning.}
Another research topic that is close to ours is graph contrastive learning, but TGS differs from it in the following two aspects: \textit{(1)} learning objective, graph contrastive learning mainly aims to learn transferable knowledge from abundant unlabeled data in an \emph{unsupervised} setting and then generalize the learned knowledge to downstream tasks. Instead, TGS works in a \emph{semi-supervised} setting, i.e., the label information is available during training. \textit{(2)} augmentation, graph contrastive learning usually requires multiple types of sophisticated augmentation to obtain different views for contrasting. However, TGS augments only positive samples by simple linear interpolation (mixup). Overall, we believe that TGS is more highly relevant to knowledge distillation than to graph contrastive learning, although we are aware that the two research topics have some aspects in common.
\newline

\noindent\textbf{Time Complexity Analysis.}
The training time complexity of the proposed TGS framework is $\mathcal{O}(|\mathcal{V}|dF+|\mathcal{E}|F)$, which is linear with respect to the number of nodes $|\mathcal{V}|$ and edges $|\mathcal{E}|$, and is in the same order of magnitude as GCNs. However, with the removal of neighborhood fetching, the inference time complexity can be reduced from $\mathcal{O}(|\mathcal{V}|dF+|\mathcal{E}|F)$ to $\mathcal{O}(|\mathcal{V}|dF)$ of MLPs. A comparison of the inference speeds of the various methods can be found in Sec.~\ref{sec:5.4} and Table.~\ref{tab:4}.

\begin{table*}[!tbp]
\begin{center}
\caption{An overview summary of the statistical characteristics of datasets.}
\label{tab:1}
\vspace{0.7em}
\resizebox{0.85\textwidth}{!}{
\begin{tabular}{lccccccc}

\toprule
\textbf{Dataset} & \texttt{Cora} & \texttt{Citeseer}  & \texttt{Amazon-Photo} & \texttt{Coauthor-CS} & \texttt{Coauthor-Phy} & \texttt{Amazon-Com} \\ \midrule
\textbf{$\#$ Nodes} & 2708 & 3327 & 7650 & 18333 & 34493 & 13752 \\
\textbf{$\#$ Edges} & 5278 & 4614 & 119081 & 81894 & 247962 & 245861 \\
\textbf{$\#$ Features} & 1433 & 3703 & 745 & 6805 & 8415 & 767 \\
\textbf{$\#$ Classes} & 7 & 6 & 8 & 15 & 5 & 10 \\
\textbf{Label Rate} & 5.2\% & 3.6\% & 2.1\% & 1.6\% & 0.3\% & 1.5\% & \\ \bottomrule

\end{tabular}}
\end{center}
\end{table*}
\begin{table*}[!tbp]
\begin{center}
\caption{Classification accuracy $\pm$ std (\%) on six real-world datasets, where the best and second results marked by \textbf{bold} and \underline{underline}, respectively. If not specifically mentioned, all relevant models adopt GCN as the backbone by default.}
\label{tab:2}
\resizebox{1.0\textwidth}{!}{
\begin{tabular}{clcccccc|c}

\toprule
\textbf{Type} & \textbf{Method} & \textbf{Cora} & \textbf{Citeseer} & \textbf{Coauthor-CS} & \textbf{Coauthor-Phy} & \textbf{Amazon-Com} & \textbf{Amazon-Photo}  & \textit{Avg. Rank} \\ \midrule
\multirow{6}{*}{MLPs / GNNs} 
 & MLP & $61.86_{\pm0.43}$ & $59.76_{\pm0.51}$ & $83.34_{\pm0.64}$ & $86.24_{\pm0.66}$ & $66.85_{\pm1.94}$ & $78.18_{\pm1.25}$ & 16.00 \\
 & GCN & $81.28_{\pm0.42}$ & $71.06_{\pm0.44}$ & $87.76_{\pm0.43}$ & $91.89_{\pm0.42}$ & $77.45_{\pm1.71}$ & $87.53_{\pm1.64}$ & 13.33 \\
 & GAT & $83.02_{\pm0.45}$ & $72.56_{\pm0.51}$ & $88.55_{\pm0.56}$ & $92.36_{\pm0.47}$ & $82.78_{\pm1.89}$ & $90.19_{\pm1.35}$ & 6.33 \\
 & GraphSAGE & $82.22_{\pm0.80}$ & $71.22_{\pm0.58}$ & $88.40_{\pm0.48}$ & $91.88_{\pm0.53}$ & $79.23_{\pm1.63}$ & $88.63_{\pm1.17}$ & 10.67 \\
 & APPNP & $83.28_{\pm0.33}$ & $71.74_{\pm0.27}$ & $88.74_{\pm0.62}$ & $92.75_{\pm0.60}$ & $81.28_{\pm1.90}$ & $89.49_{\pm1.28}$ & 6.50 \\
 & DAGNN & $\underline{84.30}_{\pm0.51}$ & $73.14_{\pm0.62}$ & $89.32_{\pm0.55}$ & $93.10_{\pm0.67}$ & $80.32_{\pm1.57}$ & $\textbf{90.72}_{\pm1.45}$ & 3.67 \\ \midrule
 
\multirow{2}{*}{GNN-to-GNN} 
 & TinyGNN & $82.79_{\pm0.57}$ & $72.67_{\pm0.72}$ & $88.72_{\pm0.42}$ & $92.20_{\pm0.67}$ & $79.22_{\pm1.69}$ & $89.24_{\pm1.24}$ & 8.67 \\ 
 & FreeKD & $83.80_{\pm0.53}$ & $73.76_{\pm0.60}$ & $89.14_{\pm0.61}$ & $92.63_{\pm0.71}$ & $80.92_{\pm1.75}$ & $89.46_{\pm1.31}$ & 5.17 \\ \midrule
 
\multirow{2}{*}{GNN Self-Distillation}  
 & GNN-SD & $81.85_{\pm0.55}$ & $71.69_{\pm0.61}$ & $87.80_{\pm0.50}$ & $92.07_{\pm0.48}$ & $77.66_{\pm1.85}$ & $87.80_{\pm1.52}$ & 11.67 \\ 
 & RDD & $83.68_{\pm0.40}$ & $73.63_{\pm0.50}$ & $\underline{89.38}_{\pm0.44}$ & $92.74_{\pm0.78}$ & $81.84_{\pm1.48}$ & $89.70_{\pm0.93}$ & 3.50 \\ \midrule

\multirow{4}{*}{GNN-to-MLP}  
 & \textcolor{table}{GNN-MLP} & \textcolor{table}{$64.53_{\pm0.42}$} & \textcolor{table}{$62.26_{\pm0.48}$} & \textcolor{table}{$81.26_{\pm0.87}$} & \textcolor{table}{$86.47_{\pm0.71}$} & \textcolor{table}{$69.25_{\pm1.75}$} & \textcolor{table}{$76.34_{\pm1.30}$} & - \\
 & GLNN & $80.85_{\pm0.60}$ & $71.21_{\pm0.80}$ & $87.81_{\pm0.53}$ & $91.83_{\pm0.60}$ & $77.96_{\pm1.70}$ & $87.98_{\pm1.36}$ & 11.29 \\
 & CPF & $83.65_{\pm0.49}$ & $72.98_{\pm0.47}$ & $89.10_{\pm0.50}$ & $92.36_{\pm0.63}$ & $80.90_{\pm1.52}$ & $89.03_{\pm1.29}$ & 6.67 \\ 
 & FF-G2M & $84.06_{\pm0.43}$ & $\underline{73.85}_{\pm0.51}$ & $88.96_{\pm0.45}$ & $92.83_{\pm0.58}$ & $81.92_{\pm1.54}$ & $89.58_{\pm1.43}$ & 3.83 \\ \midrule
 
\multirow{3}{*}{MLP Self-Distillation} 
 & \textcolor{mark}{DGI-MLP} & \textcolor{table}{$76.39_{\pm0.46}$} & \textcolor{table}{$67.83_{\pm0.51}$} & \textcolor{table}{$85.13_{\pm0.49}$} & \textcolor{table}{$89.41_{\pm0.45}$} & \textcolor{table}{$73.25_{\pm1.64}$} & \textcolor{table}{$84.60_{\pm1.60}$} & - \\
 & Graph-MLP & $81.45_{\pm0.52}$ & $72.87_{\pm0.70}$ & $88.16_{\pm0.70}$ & $91.85_{\pm0.49}$ & $77.23_{\pm1.76}$ & $87.64_{\pm1.37}$ & 11.83 \\
 & LinkDist & $76.70_{\pm0.47}$ & $65.19_{\pm0.55}$ & $87.89_{\pm0.58}$ & $92.16_{\pm0.70}$ & $76.93_{\pm1.83}$ & $87.26_{\pm1.42}$ & 13.67 \\ \midrule
 & TGS (ours) & $\textbf{84.90}_{\pm0.44}$ & $\textbf{74.08}_{\pm0.69}$ & $\textbf{89.62}_{\pm0.40}$ & $\textbf{94.96}_{\pm0.41}$ & $\underline{83.44}_{\pm2.09}$ & $90.34_{\pm0.85}$ & 1.17 \\
 & \textcolor{table}{\ \ w/o mixup} & \textcolor{table}{$83.18_{\pm0.41}$} & \textcolor{table}{$72.96_{\pm0.58}$} & \textcolor{table}{$89.18_{\pm0.61}$} & \textcolor{table}{$93.24_{\pm0.56}$} & \textcolor{table}{$82.94_{\pm1.90}$} & \textcolor{table}{$89.85_{\pm0.81}$} & - \\
 & \textcolor{table}{\ \ w/ triplet loss} & \textcolor{table}{$84.10_{\pm0.54}$} & \textcolor{table}{$73.60_{\pm0.63}$} & \textcolor{table}{$89.12_{\pm0.45}$} & \textcolor{table}{$\underline{93.86}_{\pm0.48}$} & \textcolor{table}{$\textbf{83.52}_{\pm2.10}$} & \textcolor{table}{$\underline{90.57}_{\pm0.78}$} & - \\ \bottomrule
 
\end{tabular}}
\end{center}
\end{table*}

\section{Experiments}
In this section, we evaluate TGS on six real-world datasets by answering five questions. \textbf{Q1}: How does TGS compare with SOTA graph knowledge distillation methods? \textbf{Q2}: Is TGS robust under limited labeled data and label noise? \textbf{Q3}: How does TGS compare with other general GNN models and inference acceleration methods in terms of inference speed? \textbf{Q4}: How does TGS benefit from negative samples, mixup-like augmentation, and feature/label self-distillation? \textbf{Q5}: How do two key hyperparameters, loss weight $\alpha$ and batch size $B$, influence the performance?

\subsection{Experimental Setup}
\subsubsection{Datasets} The experiments are conducted on six widely used real-world datasets, including Cora \cite{sen2008collective}, Citeseer \cite{giles1998citeseer}, Coauthor-CS, Coauthor-Physics, Amazon-Com, and Amazon-Photo \cite{shchur2018pitfalls}. For the two small-scale datasets, Cora and Citeseer, we follow the data splitting strategy in \cite{kipf2016semi}. For the four large-scale datasets, Coauthor-CS, Coauthor-Physics, Amazon-Photo, and Amazon-Computers, we follow \cite{zhang2021graph,luo2021distilling} to select 20 nodes per class to construct a training set, 500 nodes for validation, and 1000 nodes for testing. A summary of the statistical characteristics of datasets is given in Table.~\ref{tab:1}.

\subsubsection{Baselines} In this paper, we consider the following six classical baselines: Vanilla MLP, GCN \cite{kipf2016semi}, GAT \cite{velivckovic2017graph}, GraphSAGE \cite{hamilton2017inductive}, APPNP \cite{klicpera2018predict}, and DAGNN \cite{liu2020towards}. In addition, we compare TGS with four types of graph knowledge distillation methods, including (1) GNN-to-GNN KD: FreeKD \cite{feng2022freekd} and TinyGNN \cite{yan2020tinygnn}, (2) GNN-to-MLP KD: CPF \cite{yang2021extract}, GLNN \cite{zhang2021graph}, and FF-G2M \cite{wu2023extracting}, (3) GNN Self-Distillation: RDD \cite{zhang2020reliable} and GNN-SD \cite{chen2020self}, and (3) MLP Self-Distillation: Graph-MLP \cite{hu2021graph} and LinkDist \cite{luo2021distilling}. Moreover, with GCN as the base architecture, three classical inference acceleration methods are compared, including pruning with 50\% weights (P-GCN) \cite{han2015learning}, quantization from FP32 to INT8 (Q-GCN) \cite{gupta2015deep}, and neighborhood sampling with fan-out 15 (NS-GCN) \cite{chen2018fastgcn}.

\subsubsection{Hyperparameter} The hyperparameters are set the same for all datasets: Adam optimizer with learning rate $lr$ = 0.01 and weight decay $decay$ = 5e-4; Epoch $E$ = 200; Layer number $L$ = 2. The other dataset-specific hyperparameters are determined by an AutoML toolkit NNI with the hyperparameter search spaces as hidden dimension $F=\{256, 512, 1024\}$; batch size $B=\{256, 512, 1024, 4096\}$, trade-off weight $\alpha=\{0.5, 0.8, 1.0\}$. Each set of experiments is run five times with different random seeds, and the average accuracy and standard deviation are reported as metrics. Moreover, the experiments are implemented based on the standard implementation in PyTorch 1.6.0 library with Intel(R) Xeon(R) Gold 6240R @ 2.40GHz CPU and NVIDIA V100 GPU.

\begin{table*}[!htbp]
\begin{center}
\caption{Classification accuracy $\pm$ std (\%) with limited node labels, where the best results marked by \textbf{bold}.}
\label{tab:3}
\resizebox{\textwidth}{!}{
\begin{tabular}{lcccc|ccccc|cc}

\toprule
\multicolumn{2}{c}{\textbf{Dataset}} & GCN & ANNPN & DAGNN & GNN-SD & RDD & GLNN & CPF & FF-G2M & Graph-MLP & TGS (ours) \\ \midrule
\multirow{3}{*}{\textbf{Cora}} & 5 labels & $73.10_{\pm0.87}$ & $76.39_{\pm0.95}$ & $79.02_{\pm0.94}$ & $74.32_{\pm0.85}$ & $76.11_{\pm0.81}$ & $73.85_{\pm0.92}$ & $75.92_{\pm0.90}$ & $77.62_{\pm0.67}$ & $78.43_{\pm0.78}$ & $\textbf{80.22}_{\pm0.87}$ \\
 & 10 labels & $77.52_{\pm0.63}$ & $79.99_{\pm0.72}$ & $81.99_{\pm0.62}$ & $78.55_{\pm0.64}$ & $79.68_{\pm0.54}$ & $77.94_{\pm0.52}$ & $79.20_{\pm0.64}$ & $79.20_{\pm0.44}$ & $79.60_{\pm0.49}$ & $\textbf{82.80}_{\pm0.45}$ \\
 & 15 labels & $79.47_{\pm0.56}$ & $80.70_{\pm0.44}$ & $82.72_{\pm0.50}$ & $79.89_{\pm0.55}$ & $80.45_{\pm0.46}$ & $79.16_{\pm0.48}$ & $80.12_{\pm0.51}$ & $80.25_{\pm0.52}$ & $80.32_{\pm0.57}$ & $\textbf{83.40}_{\pm0.53}$ \\ \midrule
\multirow{3}{*}{\textbf{Citeseer}} 
 & 5 labels & $63.23_{\pm1.04}$ & $66.28_{\pm0.88}$ & $69.13_{\pm0.68}$ & $63.93_{\pm0.74}$ & $66.06_{\pm0.57}$ & $63.10_{\pm0.70}$ & $65.40_{\pm0.62}$ & $68.12_{\pm0.59}$ & $\textbf{69.64}_{\pm0.64}$ & $68.76_{\pm1.16}$ \\
 & 10 labels & $67.55_{\pm0.50}$ & $69.23_{\pm0.64}$ & $71.74_{\pm0.71}$ & $68.20_{\pm0.57}$ & $69.68_{\pm0.66}$ & $67.65_{\pm0.62}$ & $69.14_{\pm0.73}$ & $70.76_{\pm0.53}$ & $70.56_{\pm0.51}$ & $\textbf{72.66}_{\pm0.43}$ \\
 & 15 labels & $69.64_{\pm0.58}$ & $70.17_{\pm0.44}$ & $72.26_{\pm0.53}$ & $69.86_{\pm0.48}$ & $70.32_{\pm0.57}$ & $69.52_{\pm0.52}$ & $70.76_{\pm0.46}$ & $71.59_{\pm0.47}$ & $71.80_{\pm0.63}$ & $\textbf{73.10}_{\pm0.53}$ \\ \bottomrule
                          
\end{tabular}}
\end{center}
\end{table*}

\begin{figure*}[!tbp]
    \vspace{-1em}
	\begin{center}
		\subfigure[Robustness on the Cora dataset]{\includegraphics[width=0.315\linewidth]{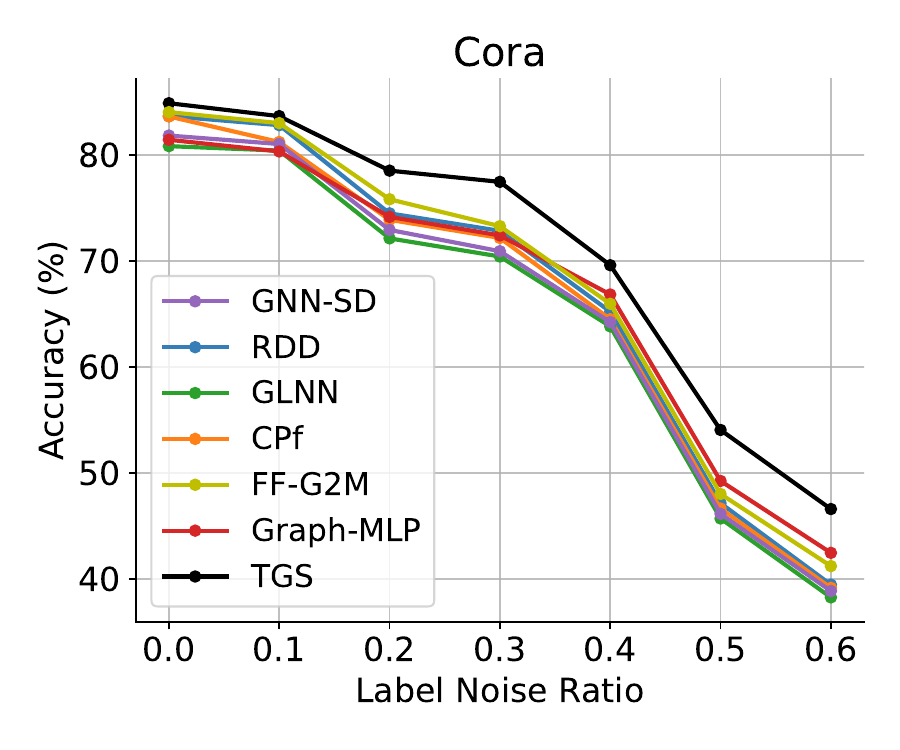}\label{fig:4a}}
		\subfigure[Robustness on the Citeseer dataset]{\includegraphics[width=0.315\linewidth]{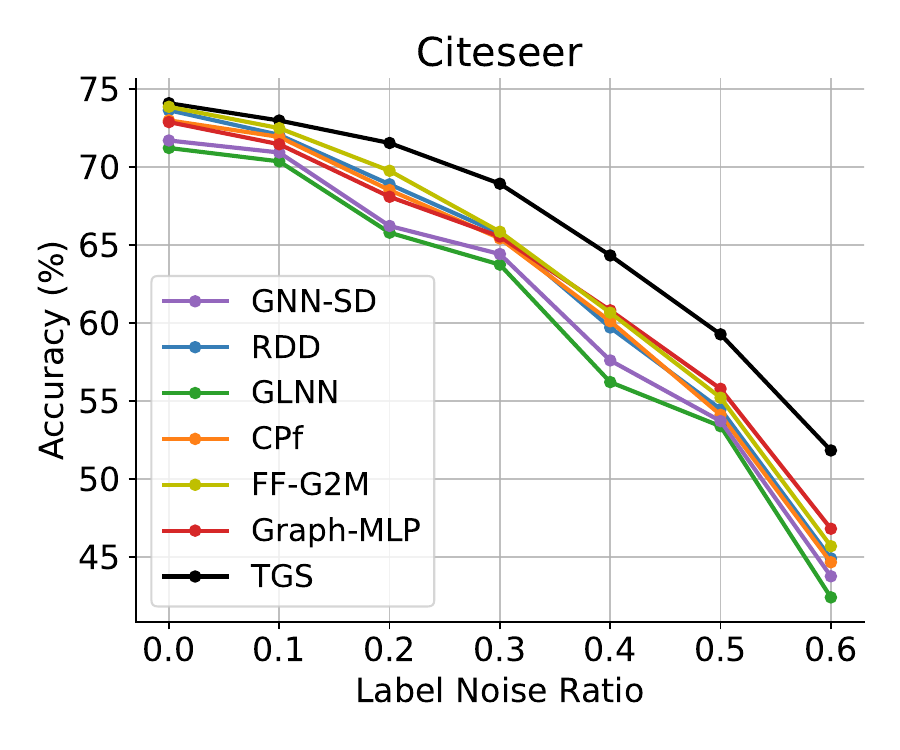}\label{fig:4b}}
		\subfigure[Inference time with different layers]{\includegraphics[width=0.35\linewidth]{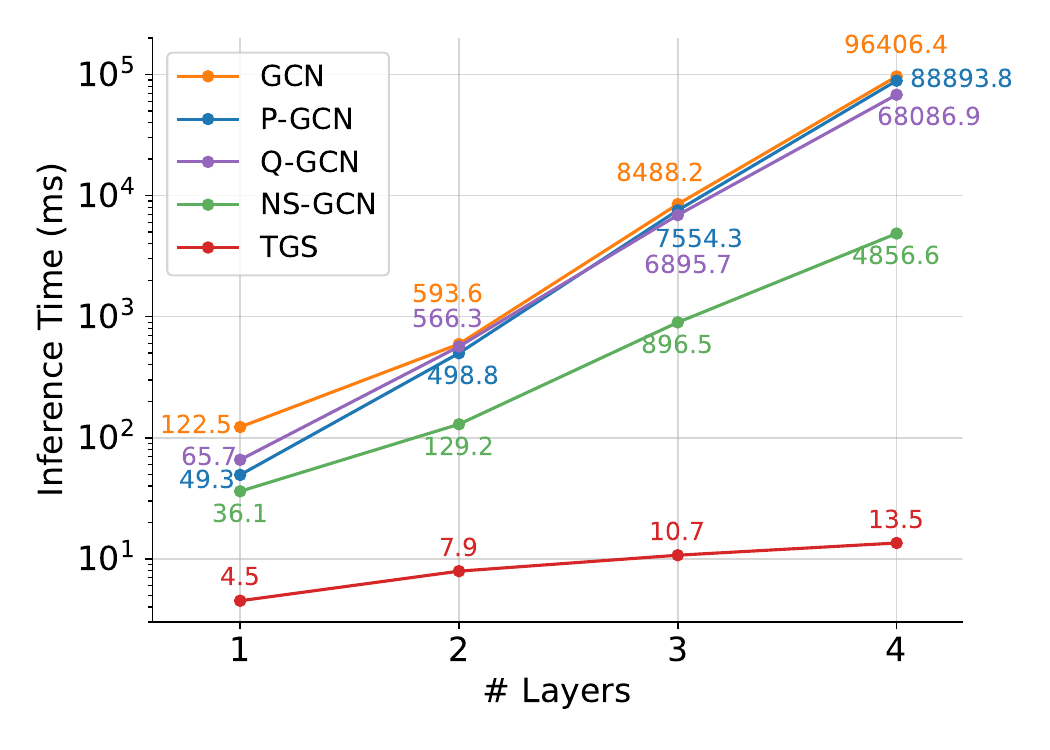}\label{fig:4c}}
	\end{center}  
    \vspace{-1em}
	\caption{(a)(b) Classification accuracy (\%) under different label noise ratios on the Cora and Citeseer datasets, respectively. (c) Inference time (ms) with different layers on the Coauthor-CS dataset.}
	\label{fig:4}
\end{figure*}

\subsection{Performance Comparison (Q1)}
To answer \textbf{Q1}, we conducted experiments on six real-world datasets with comparison with state-of-the-art baselines. From the results reported in Table.~\ref{tab:2}, we can observe that: \textit{(1)} There are some GNN-to-MLP KD and GNN Self-Distillation methods, such as RDD and FF-G2M, that can achieve comparable or even better performance than GNN-to-GNN KD, suggesting that both GNNs and teachers may not be necessary for graph knowledge distillation. (2) Considering both teacher-free and GNN-less designs, existing MLP Self-Distillation methods, such as Graph-MLP and LinkDist, lag far behind GNN-to-GNN KD and cannot even match the performance of vanilla GCNs on some datasets. \textit{(3)} Regarding classification accuracy, TGS consistently achieves the best overall performance on six datasets. For example, TGS obtains the best performance on the Coauthor-Phy dataset, and more notably, our classification accuracy is 3.13\% and 2.89\% higher than that of GLNN and GNN-SD. (4) We additionally consider two vanilla baselines: (i) GNN-MLP, which trains a GNN and then directly copies its weights to an MLP for inference, and its performance is only close to or even poorer than that of vanilla MLP; and (ii) DGI-MLP, which performs contrastive learning with DGI but uses MLP as the backbone, and its performance is far superior to that of MLP (suggesting the potential of the MLP architecture), but still inferior to that of vanilla GCN. (5) We conduct ablation studies on two important modules, namely mixup augmentation and loss function. It can be seen that mixup augmentation plays an important role in the performance. However, even with the removal of the mixup, TGS still outperforms GCN, GAT, Graph-MLP, and LinkDist, on all six datasets. However, when mixup is considered, the performance of TGS is even comparable to those state-of-the-art graph KD methods. Besides, we find that using triple loss as the objective function slightly degrades performance on some datasets, but helps improve performance on the Amazon-Com and Amazon-Photo datasets.

\subsection{Evaluation on Robustness (Q2)}
There has been some work pointing out that the performance of graph learning algorithms depends heavily on the quality and quantity of the labels. To evaluate the robustness of the TGS framework, we evaluate it with extremely limited label data and label noise on the Cora and Citeseer datasets.

\subsubsection{Performance with Limited Labels} To evaluate the effectiveness of TGS when labeled data is limited, we randomly select 5, 10, and 15 labeled samples per class for training, and the rest of the training set is considered unlabeled. From the experimental results reported in Table.~\ref{tab:3}, we can make three important observations (1) The performance of all methods drops as the number of labeled data is reduced, but that of TGS drops more slightly. (2) While GNN-to-MLP KD and GNN Self-Distillation methods perform well on clean data, as shown in Table.~\ref{tab:2}, their performance gains are reduced when labeled data is extremely limited. In contrast, the two MLP Self-Distillation methods, Graph-MLP and TGS, show a great advantage under the label-limited setting. (3) When only a limited number of labels are provided, TGS outperforms all other baselines at most label rates. For example, when trained with 5 labels per class, TGS outperforms GCN by 7.12\% and 5.53\% on the Cora and Citeseer datasets, respectively.

\subsubsection{Performance with Noisy Labels} We evaluate the robustness of TGS against label noise by injecting \textit{asymmetric} noise into class labels, where the label $i$ $(1\leq i \leq C)$ of each training sample flips independently with probability $r$ to another class, but with probability $1-r$ preserved as label $i$ \cite{yin2023omg,xia2023gnn}. The performance is reported in Fig.~\ref{fig:4a} and Fig.~\ref{fig:4b} at various noise ratios $r\in\{0\%, 10\%, 20\%, \cdots 60\%\}$. It can be seen that as noise ratio $r$ increases, the accuracy of TGS drops more slowly than other baselines, suggesting that TGS is more robust than other baselines under various noise ratios, especially under extremely high noise ratios. For example, with $r=60\%$ label noise ratio, TGS outperforms RDD and FF-G2M by 6.92\% and 6.41\% on the Citeseer dataset, respectively.

\subsection{Evaluation on Inference Speed (Q3)} \label{sec:5.4}
Commonly used inference acceleration techniques on GNNs include Pruning \cite{han2015learning}, Quantizing \cite{gupta2015deep}, and Neighborhood Sampling \cite{chen2018fastgcn}. With GCN as the base architecture, we consider its three variants: P-GCN, Q-GCN, and NS-GCN. 

\subsubsection{Comparison on Different Datasets}
With the removal of neighborhood fetching, the inference time of TGS can be reduced from $\mathcal{O}(|\mathcal{V}|dF+|\mathcal{E}|F)$ to $\mathcal{O}(|\mathcal{V}|dF)$. The inference time ($ms$) averaged over 30 sets of runs on four datasets is reported in Table.~\ref{tab:4} with the acceleration multiple w.r.t the vanilla GCN marked as ${\color[rgb]{0.4, 0.71, 0.376}green}$, where all methods use $L=2$ layers and dimension $F=16$.  From Table.~\ref{tab:4}, we can observe that: \textit{(1)} While APPNP and DAGNN improve a lot over GCN in classification accuracy, as shown in Table.~\ref{tab:3}, they suffer from more severe inference latency. (2) Pruning and quantization are not very effective on GNNs, given that data dependency in GNNs has not been well resolved. Besides, the neighborhood sampling considers but does not completely eliminate the neighborhood-fetching latency, so it infers faster than pruning and quantization, yet still lags far behind TGS which is based on MLPs. (3) TGS infers fastest across four datasets.

\begin{table*}[!htbp]
\begin{center}
\caption{Inference time ($ms$) on four datasets, where three commonly used inference acceleration methods help to speed up GCN, but still considerably slower than GSDN. Note the acceleration multiple w.r.t the vanilla GCN is marked as ${\color[rgb]{0.4, 0.71, 0.376}green}$.}
\label{tab:4}
\resizebox{0.8\textwidth}{!}{
\begin{tabular}{lccccccc}

\toprule
\textbf{Method} & GCN & APPNP & DAGNN & P-GCN & Q-GCN & NS-GCN & TGS (ours) \\ \midrule
\textbf{Cora} & 402.7 & 731.8 & 647.6 & 372.9 (${\color[rgb]{0.4, 0.71, 0.376}1.08\times}$) & 383.5 (${\color[rgb]{0.4, 0.71, 0.376}1.05\times}$) & 97.7 (${\color[rgb]{0.4, 0.71, 0.376}4.12\times}$) & 4.6 (${\color[rgb]{0.4, 0.71, 0.376}87.54\times}$) \\
\textbf{Citeseer} & 383.4 & 700.5 & 727.5 & 316.9 (${\color[rgb]{0.4, 0.71, 0.376}1.21\times}$) & 361.4 (${\color[rgb]{0.4, 0.71, 0.376}1.06\times}$) & 105.6 (${\color[rgb]{0.4, 0.71, 0.376}3.63\times}$) & 4.3 (${\color[rgb]{0.4, 0.71, 0.376}89.16\times}$) \\
\textbf{Coauthor-CS} & 593.6 & 1099.1 & 672.4 & 498.8 (${\color[rgb]{0.4, 0.71, 0.376}1.19\times}$) & 566.3 (${\color[rgb]{0.4, 0.71, 0.376}1.05\times}$) & 129.2 (${\color[rgb]{0.4, 0.71, 0.376}4.59\times}$) & 7.9 (${\color[rgb]{0.4, 0.71, 0.376}75.14\times}$) \\
\textbf{Coauthor-Phy} & 1067.6 & 1467.4 & 708.6 & 922.3 (${\color[rgb]{0.4, 0.71, 0.376}1.16\times}$) & 997.8 (${\color[rgb]{0.4, 0.71, 0.376}1.07\times}$) & 269.3 (${\color[rgb]{0.4, 0.71, 0.376}3.96\times}$) & 12.8 (${\color[rgb]{0.4, 0.71, 0.376}83.41\times}$) \\ \bottomrule

\end{tabular}}
\end{center}
\end{table*}

\subsubsection{Comparison with Different Layers}
To infer a single node with a $L$-layer GNN on a graph with average node degree $R$, it requires fetching $\mathcal{O}(R^L)$ nodes. To compare the sensitivity of TGS to layer depth with other inference acceleration methods, we report their inference time ($ms$ in log-scale) at different layer depths on the Coauthor-CS dataset in Fig.~\ref{fig:4c}. The inference time of TGS only increases \textbf{linearly} with the layer depth, but that of other baselines increases \textbf{exponentially}. Moreover, the speed gains of GNN pruning and quantization are reduced as the layer depth increases, and they approach the vanilla GCN when the layer depth is 4. In contrast, at larger layer depths, the speed gain of the neighbor sampling gets enlarged compared to the vanilla GCN. This demonstrates that \textit{neighborhood fetching is one major source of inference latency in GNNs}, and the linear complexity of TGS has a great advantage, especially when GNNs become deeper.

\begin{figure*}[!tbp]
    \vspace{-1em}
	\begin{center}
        \subfigure[Ablation Study]{\includegraphics[width=0.33\linewidth]{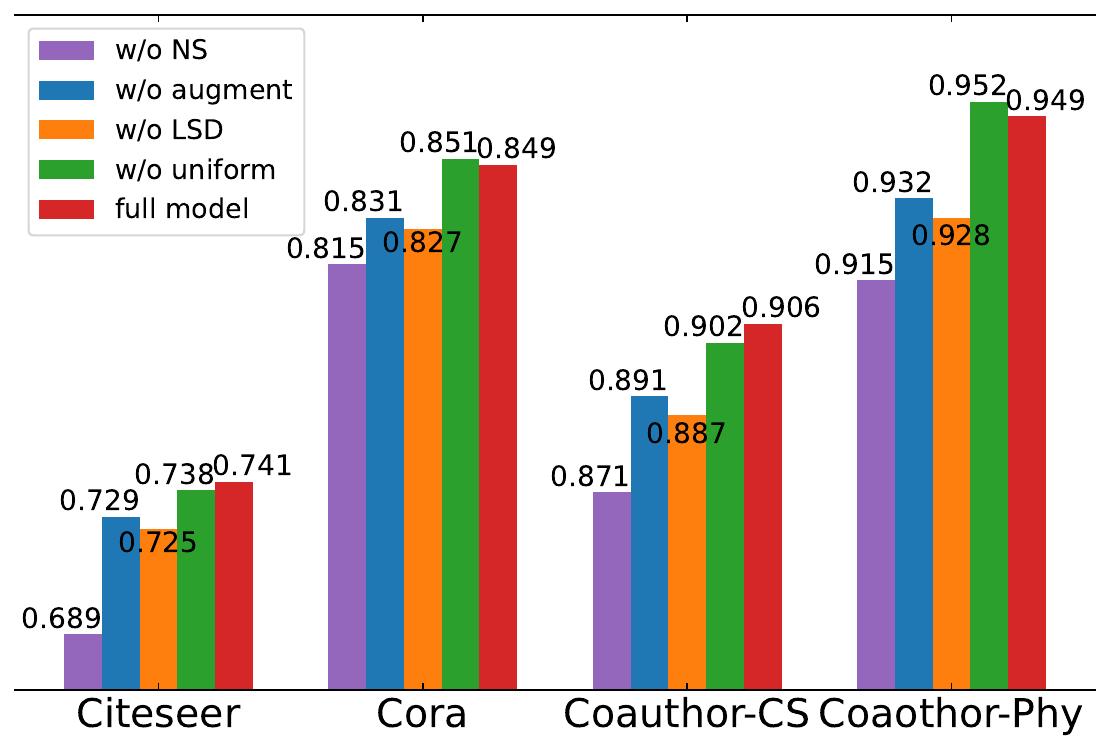}\label{fig:5a}}
		\subfigure[Learning curves]{\includegraphics[width=0.32\linewidth]{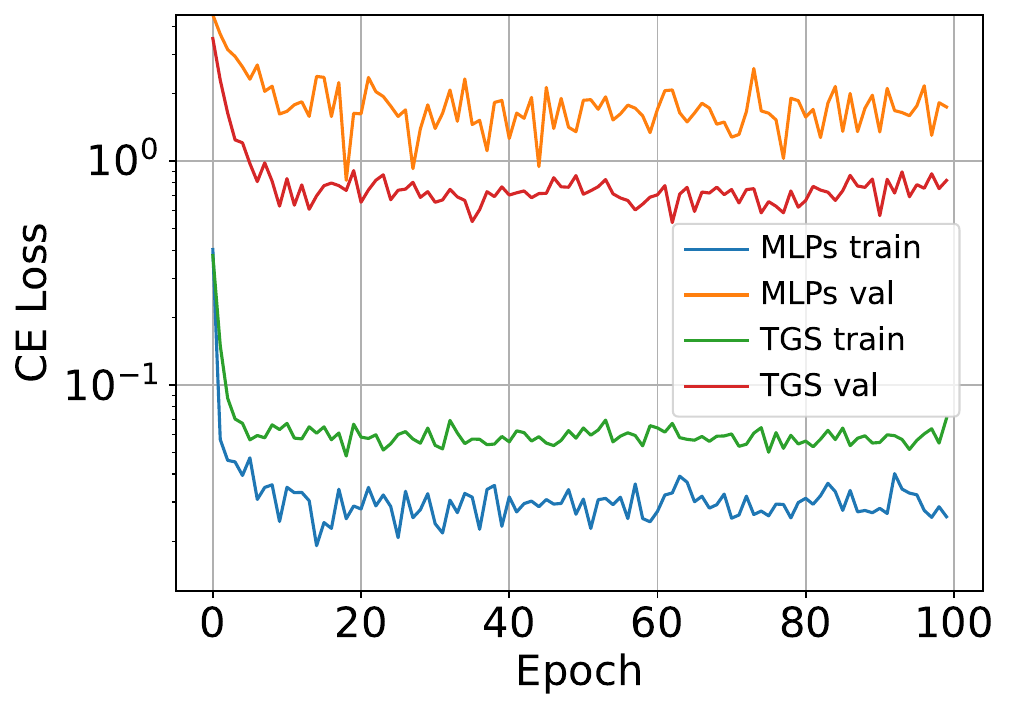}\label{fig:5b}}
		\subfigure[Mean cosine similarity curves]{\includegraphics[width=0.32\linewidth]{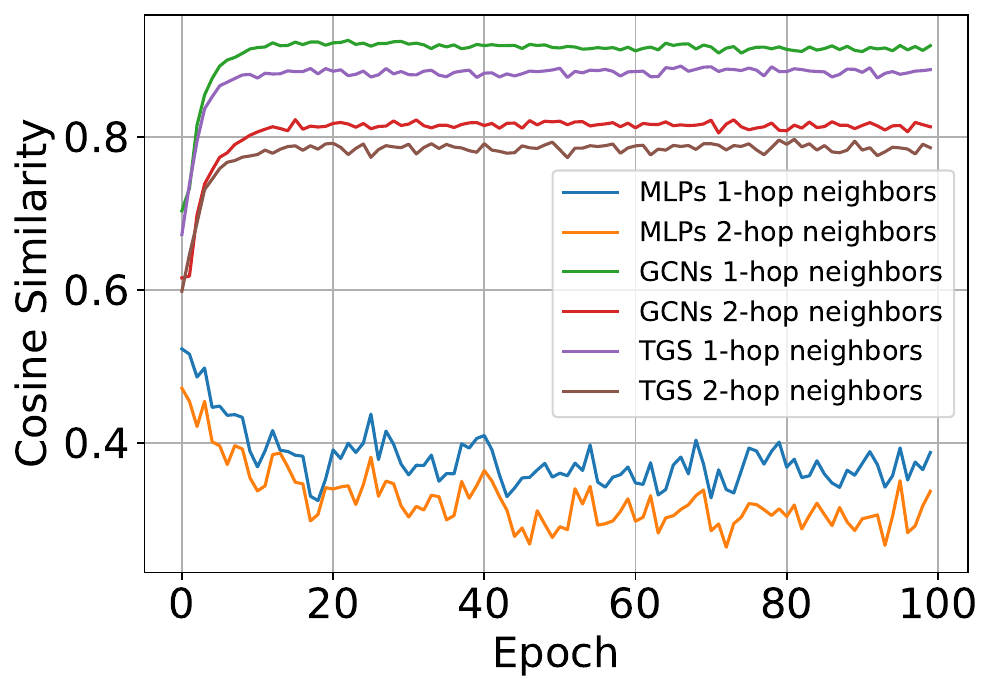}\label{fig:5c}}
	\end{center}
    \vspace{-1em}
	\caption{(a) Ablation study on four key model components. (b) Learning curves of MLPs and TGS on the Cora dataset, showing that self-distillation helps to regularize the training. The \emph{logarithmized} vertical coordinate is the cross-entropy loss between the predicted and ground-truth labels on the training or validation set, respectively. (c) Mean cosine similarity curves of MLPs, GCNs, and TGS between the target node with 1-hop and 2-hop neighbors on the Cora dataset.}
	\label{fig:5}
\end{figure*}

\subsection{Ablation Study (Q4)} \label{sec:5.5}
\subsubsection{Component Analysis}
To evaluate the effectiveness of negative samples in Eq.~(\ref{equ:4}), mixup-like augmentation in Eq.~(\ref{equ:3}), and label self-distillation in Eq.~(\ref{equ:8}), we conducted four sets of experiments: the model without (A) Negative Samples (\textit{w/o} NS); (B) mixup-like augmentation (\textit{w/o} augment); (C) Label Self-Distillation (\textit{w/o} LSD); and (D) the full model. Besides, to evaluate the impact of the negative distribution, we take the nodal degree $d_i$ as a prior and preset $P_k(v_i)=\frac{d_i}{|\mathcal{E}|}$ in place of the default uniform distribution in this paper, denoted as (E) \textit{w/o} uniform. After analyzing the results in Fig.~\ref{fig:5a}, we can conclude that: (1) Negative Samples and mixup-like augmentation contribute to improving classification performance. More importantly, applying them together can further improve performance on top of each. (2) Label self-distillation helps to improve performance on top of the stand-alone feature self-distillation. (3) Even without considering any graph prior, presetting $P_k(\cdot)$ as uniform distribution is sufficient to achieve comparable performance, so this paper defaults to the simplest uniform distribution without considering complex prior-based distributions.

\subsubsection{How TGS Benefit from Self-Distillation}
Previous attempts have shown that there do exist optimal MLP parameters enabling its performance to be competitive with GNNs, but it is hard to learn such parameters by only cross-entropy loss \cite{hu2021graph,luo2021distilling}. The proposed TGS helps to solve this problem with two potential advantages: (1) alleviating overfitting, and (2) introducing graph topology \cite{zhang2021graph}.

Firstly, we plot the training curves (with log-scale vertical coordinate) of TGS and MLPs on the Cora dataset in Fig.~\ref{fig:5b}, from which we observe that the gap between training and validation loss is smaller for TGS than MLPs, which indicates that TGS helps to alleviate the overfitting trend of MLPs. Secondly, we conjecture that the absence of inductive bias, e.g., graph topology, is one of the major reasons why MLP is inferior to GNN in inference accuracy. To illustrate it, we plot in Fig.~\ref{fig:5c} the average cosine similarity of nodes with their 1-hop and 2-hop neighbors for MLPs, GCNs, and TGS on the Cora dataset. It can be seen that the average similarity with 1-hop neighbors is always higher than that with 2-hop neighbors throughout the training process for MLPs, GCNs, and TGS. More importantly, the average similarity of GCNs and TGS gradually increases with training, while that of MLPs gradually decreases, which indicates that TGS has introduced graph topology as an inductive bias (as GCN has done), while MLP does not. As a result, our TGS enjoys the benefits of topology-awareness in training but without neighborhood-fetching latency in inference.

\subsection{Hyperparameter Sensitivity Analysis (Q5)}
\begin{figure}[!htbp]
	\begin{center}
		\includegraphics[width=0.50\linewidth]{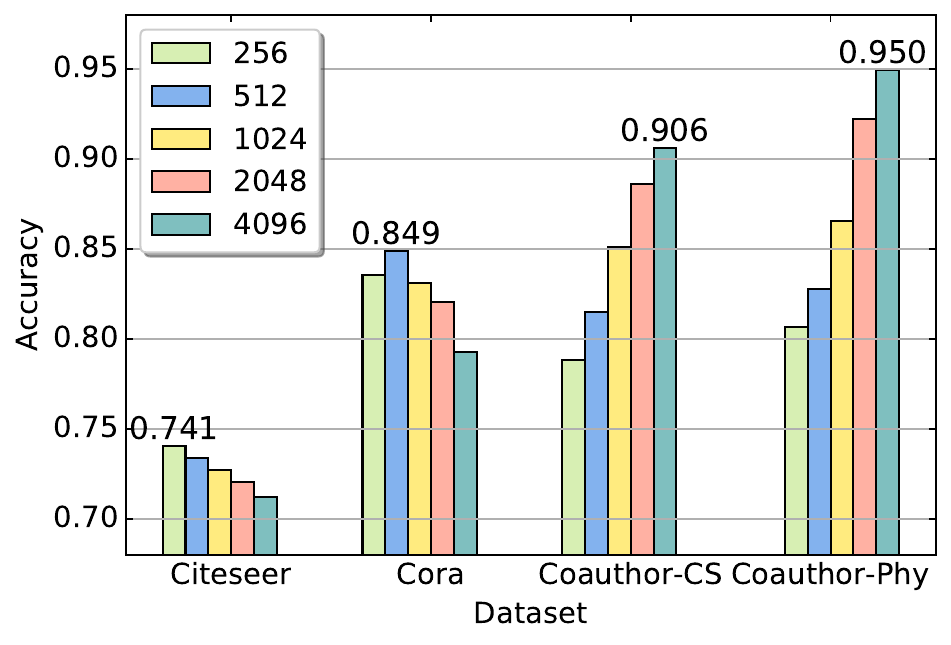}
		\includegraphics[width=0.49\linewidth]{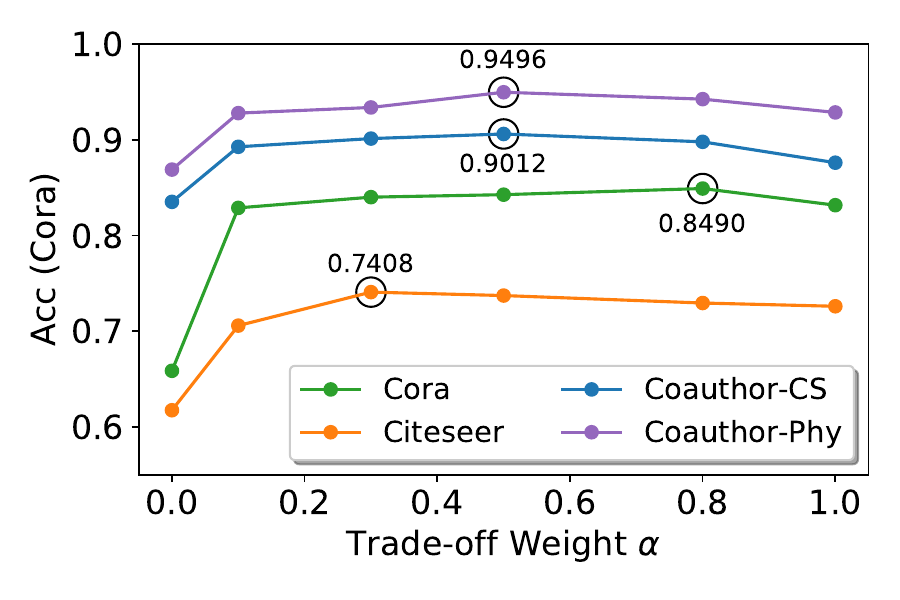}
	\end{center}
	\caption{Hyperparameter sensitivity analysis on the batch size $B$ (\textit{Left}) and trade-off weight $\alpha$ (\textit{Right}) on four datasets.}
	\label{fig:6}
\end{figure}

To answer \textbf{Q5}, we evaluate the hyperparameter sensitivity w.r.t two key hyperparameters: trade-off weight $\alpha\in\{0.0, 0.1, 0.3, 0.5, 0.8, 1.0\}$ and batch size $B\in\{256, 512, 1024, 2048, 4096\}$ in Fig.~\ref{fig:6}, from which we can make two key observations that (1) batch size $B$ is a dataset-specific hyperparameter. For simple graphs with few nodes and edges, such as Cora, a small batch size, $B=256$, can yield fairly good performance. However, for large-scale graphs with more nodes and edges, such as Coauthor-Phy, the model performance usually improves with the increase of batch size $B$. (2) When $\alpha$ is set to 0, i.e., the feature-level self-distillation is completely removed, the performance of TGS degrades to be close to that of MLPs. In contrast, when $\alpha$ takes a non-zero value, the performance of TGS improves as $\alpha$ increases. However, when $\alpha$ becomes too large, it weakens the benefit of label information, yielding lower performance improvements. In practice, we can usually determine $B$ and $\alpha$ by selecting the model with the highest accuracy on the validation set through the grid search.

\section{Conclusion}
Motivated by the complementary strengths and weaknesses of GNNs and MLPs, we propose \emph{\underline{T}eacher-Free \underline{G}raph \underline{S}elf-Distillation} (TGS) framework that does not require both teacher models and GNNs during training and inference. The proposed TGS framework is purely based on MLPs, where structural information is only implicitly used to guide dual knowledge self-distillation between the target node and its neighborhood. More importantly, we study TGS comprehensively by investigating how they benefit from neighborhood self-distillation and how they are different from existing works. Extensive experiments have shown the advantages of TGS over existing methods in terms of both inference accuracy and inference efficiency. Despite the great progress, limitations still exist and a major concern is that TGS is based on the neighborhood smoothing assumption (a common assumption adopted by various models such as GCN, GAT, etc.), so how to extend TGS to heterophily graphs may be a promising direction for future work.

\section{Acknowledgement}
This work was supported by National Key R\&D Program of China (No. 2022ZD0115100), National Natural Science Foundation of China Project (No. U21A20427), and Project (No. WU2022A009) from the Center of Synthetic Biology and Integrated Bioengineering of Westlake University.

\bibliographystyle{IEEEtran}
\bibliography{Bibliography-File}

\end{document}